\documentclass{article}
\usepackage{amsmath,epsfig,amsmath,bm,bbding,amsfonts,graphicx,times,subfigure}
\usepackage[preprint]{spconfa4}
\usepackage{xcolor}
\usepackage{cite}

\let\OLDthebibliography\thebibliography
\renewcommand\thebibliography[1]{
  \OLDthebibliography{#1}
  \setlength{\parskip}{0pt}
  \setlength{\itemsep}{0pt plus 0.3ex}
}

\begin{document}\sloppy

% Example definitions.
% --------------------
\def\x{{\mathbf x}}
\def\L{{\cal L}}

% Title.
% ------
\title{DIVERSE INSTANCE DISCOVERY: VISION-TRANSFORMER FOR INSTANCE-AWARE MULTI-LABEL IMAGE RECOGNITION}
%
% Address.
% ---------------
\name{Yunqing Hu$^{1,2}$, Xuan Jin$^{2}$, Yin Zhang$^{1\ast}$\thanks{$^{\ast}$Corresponding author: Yin Zhang.}, Haiwen Hong$^{1,2}$, Jingfeng Zhang$^{1,2}$, Feihu Yan$^{1}$, Yuan He$^{2}$, Hui Xue$^{2}$}
\address{$^{1}$College of Computer Science and Technology, Zhejiang University, Hangzhou, China, \\$^{2}$Alibaba Group, Hangzhou, China}
% \address{$^{\ast}$First author address and e-mail; $^{\dagger}$Second author address and e-mail ; (...) and  $^{\ddagger}$Last author address and e-mail.}

\maketitle
\begin{abstract}
Previous works on multi-label image recognition (MLIR) usually use CNNs as a starting point for research. In this paper, we take pure Vision Transformer (ViT) as the research base and make full use of the advantages of Transformer with long-range dependency modeling to circumvent the disadvantages of CNNs limited to local receptive field. However, for multi-label images containing multiple objects from different categories, scales, and spatial relations, it is not optimal to use global information alone. Our goal is to leverage ViT's patch tokens and self-attention mechanism to mine rich instances in multi-label images, named diverse instance discovery (DiD). To this end, we propose a semantic category-aware module and a spatial relationship-aware module, respectively, and then combine the two by a re-constraint strategy to obtain instance-aware attention maps. Finally, we propose a weakly supervised object localization-based approach to extract multi-scale local features, to form a multi-view pipeline. Our method requires only weakly supervised information at the label level, no additional knowledge injection or other strongly supervised information is required. Experiments on three benchmark datasets show that our method significantly outperforms previous works and achieves state-of-the-art results under fair experimental comparisons.
\end{abstract}
\begin{keywords}
Multi-label image recognition, vision transformer, instance-aware
\end{keywords}
\section{Introduction}
\label{sec:intro}

\noindent Multi-label image recognition (MLIR) is a practical and challenging computer vision task. Compared with single-label image recognition, the key and challenge of MLIR is the recognition of objects with different categories, scales, and spatial locations in an image. There are many real-world application scenarios for MLIR, such as Autonomous Driving \cite{driving}, multimodal analysis \cite{HatefulMeme}, and human attribute recognition \cite{Human}, etc. In the early studies of MLIR, it was found that although CNNs \cite{ResNet, spanet} have strong global representation capabilities for images, however, in the face of MLIR tasks, CNN models pre-trained on single-label datasets, such as ImageNet \cite{ImageNet}, are still not optimal. 

% We summarize the characteristics of existing typical methods in Table 1. 
Some methods, e.g., HCP \cite{HCP} borrow the idea of object detection and introduce region proposals in MLIR. 
% However, this kind of methods require manually labeled bounding box annotations, and need to generate and process a large number of region proposals that do not necessarily contain the real targets.
The other kind of methods requires the use of label dependencies, including static co-occurrence statistics between labels and word vector embedding of labels \cite{icme1}.
For example, ML-GCN \cite{ML-GCN} and SSGRL \cite{SSGRL} construct co-occurrence relationships between labels in the datasets and introduce GCN at the same time. 
Meanwhile, there are some recent works \cite{GM-MLIC, joint}, which introduce large pre-trained detection models, such as Faster-RCNN \cite{faster-rcnn} to extract the categories as well as the spatial relationships of multiple objects in an image, while combining with graph neural networks such as GCN \cite{gcn}.
We refer to such methods as strong prior knowledge injection methods. However, the two above kinds of methods face the problem of unstable static statistical relationships and poor generalization performance.

It is consistent with human visual cognitive habits to locate and then recognize many region proposals containing instances in the global space of a given image. However, we notice that few methods explore the rapid localization and recognition of rich instances in images. 
% Instead of using GCNs to painstakingly represent the potential higher-order relationships between labels, it is better to use hard attention to explicitly strengthen the relationships between local instances and labels.
Last but not least, due to the high speed of Vision Transformer (ViT) \cite{ViT} development, some Transformer-based MLIR works \cite{C-Trans,csra} emerged, however, the backbones of these methods are still CNNs model, which cannot better exploit the Transformer's ability to model global dependencies among features. 

Based on the above analysis, we raise a question: \emph{Could we further improve the instance-aware model's ability to automatically perceive various categories, scales, and locations just purely based on the ViT itself rather than CNN, while considering local relationships at the spatial scale? }
%thus enables the model to be instance-aware?}

To this end, we propose Diverse instance discovery: Vision-Transformer for instance-aware Multi-label Image Recognition. The core of our method lies in obtaining instance-aware attention maps, which are used for prediction proposals generation through a dynamic instance localization method. The prediction proposals allow local instance regions to be selected on the raw image for integrated multi-label learning. To achieve the purpose of instance-awareness, we divide the whole process into two steps, firstly, we linearly map the high-dimensional patch tokens of ViT to the low-dimensional feature maps with category semantic awareness. Secondly, we propose a re-constraint strategy to enhance the spatial relation constraints to the above low-dimensional features through the self-attention mechanism that comes with ViT, to obtain the instance-aware attention maps. 
As for dynamic instance localization, we propose a weakly supervised object localization-based approach to obtain local instance region features, which together with global features form a multi-view pipeline.
The contributions of this paper are as follows: (1) To the best of our knowledge, we are the first to explore the work of pure ViT on MLIR, effectively exploiting the long-range dependencies among patch tokens, without using CNNs as backbone at all as in previous works. (2) We propose to effectively discover different instances through instance-aware attention maps, providing an effective perspective to exploit the semantic category information and spatial relationship information extracted by ViT for MLIR. (3) Experimental results on three widely used benchmarks (MS-COCO, VOC2007, and VOC2012) demonstrate the superiority of our proposed method against state-of-the-art works.

\section{Proposed Method}
\subsection{Vision Transformer}
The main structure of ViT blocks consists of a stack of L Transformer's standard encoder. Each block consists of a multi-head self-attention (MSA) and a feed-forward network (FFN), which consists of two fully connected layers. The output of the k-th layer can be expressed as :
\begin{equation}
% \scriptsize
    y_k = LN(x_{k-1}+MSA(x_{k-1}))
\end{equation}
\begin{equation}
% \scriptsize
    x_k = LN(y_k+FFN(y_k))
\end{equation}
For an image $x$, $x \in \mathbb{R}^{H \times W \times 3}$, where $H$ and $W$ are the 
height and width of the raw image respectively, and 3 is the number of channels of the raw RGB image. ViT first performs image tokenization: cut $x$ into $N$ patches 
with a certain patch size $P$, where $N = H \times W /P^2$. Then ViT flattens each patch and performs a linear transformation to a specific dimension $D$ to obtain the patch token: $x_p \in \mathbb{R}^{N \times D}, p \in {1,2,...,D}$. 
Similar to BERT, ViT randomly initializes the class token for final classification, which will be merged with the patch tokens and be fed into the subsequent transformer. In addition, since the patch tokens fed into the subsequent transformer are position-agnostic, while image processing depends on the position information of each pixel, a learnable position embedding is added to each token in an element-wise manner.

\subsection{Overview of DiD}
Figure 1 depicts our overall framework. First, the input image $\mathbf{x} \in \mathbb{R}^{H \times W \times 3}$, is processed by  the L-layer Encoder of ViT to obtain the output patch tokens $\mathbf{x}_p \in \mathbb{R}^{N \times D}$ of the last encoder layer, where $N=H \times W/P^2$, $P$ is the patch size. We refer to $\mathbf{x}_p$ as the global features.
In the second step, the patch tokens $\mathbf{x}_p$ will be mapped to features related to semantic categories by the semantic category-aware module.
In the third step, through the spatial relationship-aware module, we get the self-attentive weights of ViT, which will be combined with the features related to semantic categories by the re-constraint strategy. 
Finally, we extract the local instance region features from the raw image by the dynamic instance localization module, to form a multi-scale feature together with the global features,  as well as a multi-view pipeline.

\subsubsection{Semantic category-aware module}
ViT uses a learnable class token as the input to the classifier after the last encoder layer. However, the class token is a one-dimensional vector containing only the global information needed for classification. If we want to generate the semantic category-aware attention maps of two-dimensional, then we must necessarily pick up the originally discarded patch tokens $\mathbf{x}_p \in \mathbb{R}^{N \times D}$.
Inspired by MCAR \cite{mcar}, we first reshape $\mathbf{x}_p$ to get $\mathbf{x}' \in \mathbb{R}^{D \times w \times h}$, where $h = w = N^{1/2}$, and we use $\mathbf{x}_d'$ to denote the $d$-th feature map, $d=1,2,...,D$ then we apply a $1 \times 1$ convolutional layer to the $\mathbf{x}_d'$:
\begin{equation}
% \scriptsize
    A_c = \sum_{d}\mathbf{x}_d'*k_{c,d}
\end{equation}
where $k \in \mathbb{R}^{C \times D \times 1 \times 1}$, $C$ is the number of categories, $\ast$ is the convolution operation, $k_{c,d}$ is a $1 \times 1$ kernel map indexed by $c$ and $d$.
Then we transform the reduced-dimensional feature maps $\mathbf{A} \in \mathbb{R}^{h\times w \times C}$ into a vector $B \in \mathbb{R}^{C}$, by average pooling operation. $B$ is the semantic category-aware feature vector, which will be continuously optimized by the standard multi-label classification Loss. 
It is a complete training process up to here, however, the preliminary experiments (as shown in the supplementary material) tell us that 
%the classification results obtained after dimensionality reduction of the patch token 
using patch tokens in this way are not better than using the class token. Thus maybe there is more potential for patch tokens to be explored.
Transforming $\mathbf{A}$ into $B$ after the average pooling operation can be considered as a class-specific global spatial attention mechanism. We speculate that the reason for the poor results of training $B$ directly for classification is the lack of spatial relationship enhancement for specific categories. The magnitude of $B$, i.e., the prediction confidence, is equivalent to the probability of the possible occurrence of the category.
Inspired by MCAR \cite{mcar}, we sort $B$ in descending order by size and take the topN feature maps among them for the next step of spatial relationship-aware enhancement and dynamic instance localization.

\subsubsection{Spatial relationship-aware module}
Inspired by attention flow \cite{flow}
% and RAMS-Trans, 
we propose to use ViT's self-attention weights to add constraints on spatial relationships to the feature maps generated in the previous section. Specifically, we first take out the attention weights of each layer of transformer:
\begin{equation}
% \scriptsize
    \bm{W_{h}^{l}} = softmax(\frac{\bm{Q}\bm{K}^T}{\sqrt{C/H}})
\end{equation}
where $\bm{W_{h}^{l}} \in \mathbb{R}^{(N+1) \times (N+1)}$, $h=1,2,...,H$ and $l=1,2,...,L$, $H$ and $L$ are the number of heads and layers. $\bm{Q}$, $\bm{K}$ are Query and Key vectors of all patches respectively. The normalized attention weights matrix $G_l$ of layer $l$ is defined as follows:
\begin{equation}
% \scriptsize
    \bm{G_l} = normc( \frac{1}{H}\sum_{h={1}}^{H}\bm{W_{h}^{l}}+\bm{E})
\end{equation}
where $\bm{E}$ is the diagonal matrix, $normc(\bm{M})$ normalizes each column of the matrix $\bm{M}$ to a probability vector by dividing each element in the column by the sum of all elements in that column. Then we integrate the attention weights for each of the previous layers by recursive matrix multiplication: $V = \prod_{l=1}^{L}{\bm{G_l}}$,
% \begin{equation}
% % \scriptsize
%     V = \prod_{l=1}^{L}{\bm{G_l}}
% \end{equation}
where $V \in \mathbb{R}^{(N+1) \times (N+1)}$. We take out the first row of the first dimension of $V$ and all the values after the first column of the second dimension of $V$ to form $V^{'}$, $V^{'} \in \mathbb{R}^{1 \times N}$,  which is the attention weights corresponding to the class token. Then we reshape $V'$ to $N^{1/2} \times N^{1/2}$. Through Equation (4) we know that $V'$ represents the mutual relationship between class token and all patch tokens, which means $V'$ represents its degree of attention to each patch token to some extent. And because the patch tokens represent the corresponding spatial relationship between themself and the pixel blocks of the raw image in ViT, here we use $V'$ as a constraint on the spatial relationships of all objects in the image.

\begin{figure}[ht]
\centering
\includegraphics[width=1.1\columnwidth]{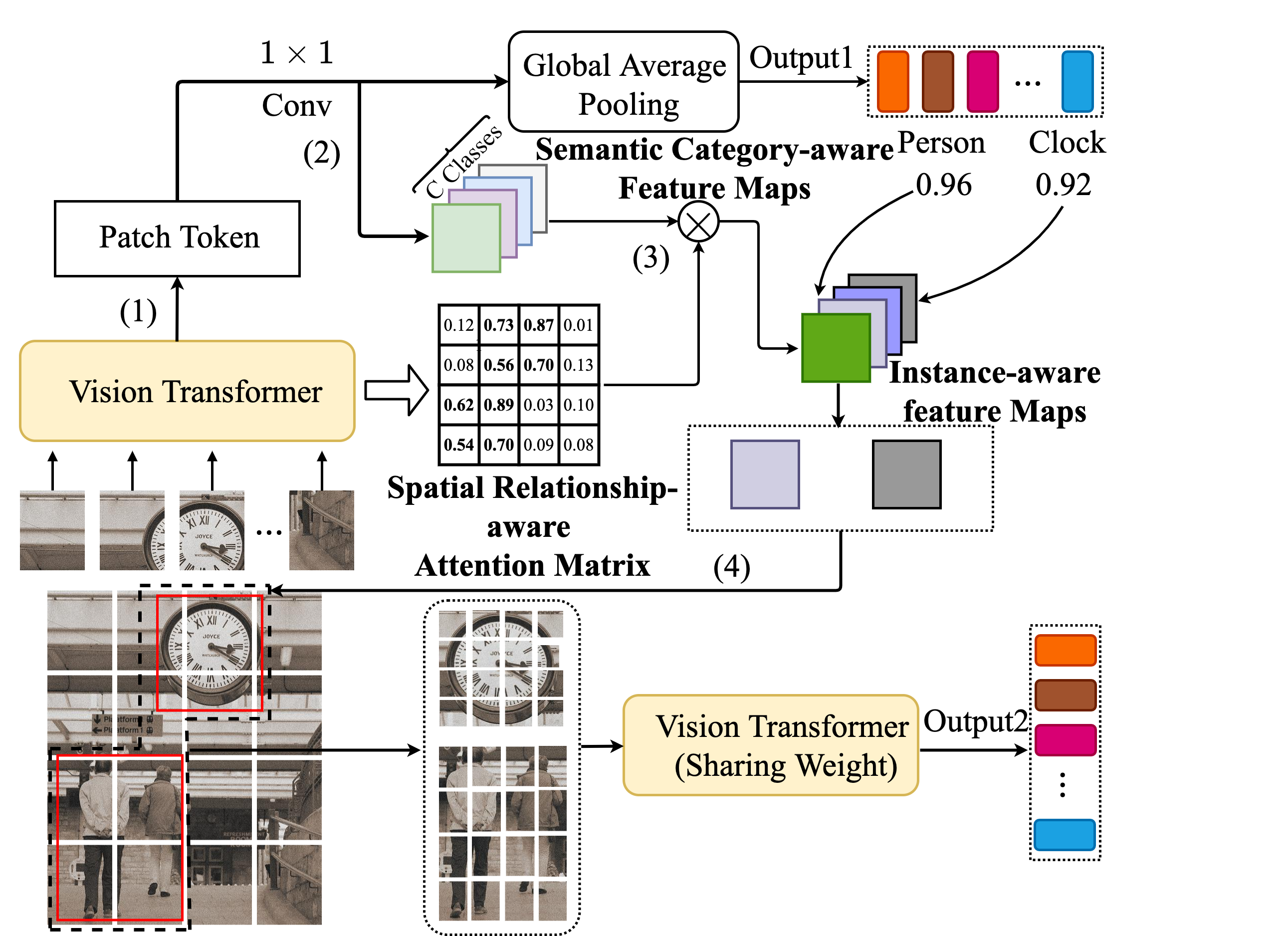} % Reduce the figure size so that it is slightly narrower than the column. Don't use precise values for figure width.This setup will avoid overfull boxes.
\caption{
The overall architecture of our DiD model. (1) The raw image passes through ViT to extract features and get the patch tokens. (2) The patch tokens are processed by a $1\times1$ convolution to obtain the semantic category-aware feature maps. (3) The self-attention weights are processed to form the spatial relationship-aware matrix, and the instance-aware feature maps are obtained by combining the re-constraint strategy with the feature maps obtained in the second step. (4) The prediction proposals are obtained from the feature maps obtained in the third step, which guide the model to be trained again.}
\end{figure}

\subsubsection{Instance-aware}
After obtaining semantic category-awareness and spatial relation-awareness from the above two sections, respectively, we will fuse spatial relations into the semantic category-aware feature maps to obtain the instance-aware attention maps, which we call re-constraint. To this end, we design three re-constraint strategies. The first is based on the Hadamard product:
\begin{equation}
% \scriptsize
    S = V'\otimes A
\end{equation}
The second is a strategy based on element-wise summation:
\begin{equation}
% \scriptsize
    S = V'\oplus A
\end{equation}
The third is the strategy of not integrating spatial relationships:
\begin{equation}
% \scriptsize
    S = A
\end{equation}
We want to enhance the constraints of spatial relations on the original semantic category-aware feature maps through re-constraint strategies. We will specifically compare the advantages and disadvantages of these re-constraint strategies in the experimental section.

\subsubsection{Dynamic instance localization}
To obtain the prediction proposals of TopN categories from the above obtained instance-aware attention maps $A$, inspired by weakly supervised object localization \cite{CAM}, we propose dynamic instance localization.
Specifically, first we upsample $\mathbf{A}$ to the same size as the raw image $x$ by the bilinear interpolation method to obtain $A'$ for pixel-level alignment of semantic and spatial relationships. In this case, the value of each position $(x, y)$ in $A'$ represents the probability that it belongs to category $i$.
To generate the predicted bounding box from $A'$, we use $\lambda$, $0<\lambda <1$ as the adjustment coefficient to dynamically adjust the segmentation area for each image.
We use $100 \times \lambda$ \% of the maximum value of $A'$ as the threshold to split the foreground and background areas of $A'$, and we consider the part larger than the threshold as the foreground area and the part smaller than the threshold as the background area.
Then we take out the predicted bounding box of the largest connected component that covers the foreground segmentation region.
It should be noted that the key to the classification problem lies in the extraction of discriminable regions, so in practice, we need to appropriately turn up the $\lambda$ value to achieve the best results, as analyzed in our experimental section.

\section{Experiments}
In this section, to validate the effectiveness of our proposed method and to compare with other state-of-the-art methods, we conduct multiple sets of experiments and present specific experimental details and results for 3 mainstream MILR datasets, including MS-COCO \cite{MS-COCO}, PASCAL-VOC2007 \cite{VOC2007} and PASCAL-VOC2012 \cite{VOC2012}. We perform an exhaustive analysis of the experiments for the method in this paper, to investigate more deeply the impact of the method acting on MILR.

\begin{table*}[tb]
\footnotesize
\centering
\caption{Comparison of our DiD and other state-of-the-art methods on MS-COCO dataset. The best results are marked as bold.}
\begin{tabular}{c|l|clllll|clllll}
\hline
& \multicolumn{1}{l}{}  & \multicolumn{6}{c}{All}  & \multicolumn{6}{c}{Top3}   \\ \hline
Methods  & mAP  & CP & CR. & CF1 & OP  & OR  & OF1  & CP & CR  & CF1    & OP    & OR  & OF1   \\ \hline\hline
% ResNet-101 \cite{ResNet}  & 79.4  & 83.4  & 66.6  & 74.0  & 86.8  & 71.1 & 78.2  
%                     & 86.2  & 59.7  & 70.6  & 90.5  & 63.7  & 74.8  \\
                    
ML-GCN \cite{ML-GCN}      & 83.0  & 85.1  & 72.0  & 78.0  & 85.8  & 75.4  & 80.3                          & 89.2  & 64.1  & 74.6  & 90.5  & 66.5  & 76.7  \\

MS-CMA \cite{cross-modality}      & 83.8  & 82.9  & 74.4  & 78.4  & 84.4  & 77.9  & 81.0  
                    & 88.2  & 65.0  & 74.9  & 90.2  & 67.4  & 77.1  \\

% KSSNet \cite{kd}     & 83.7  & 84.6  & 73.2  & 77.2  & 87.8  & 76.2  & 81.5                          & -     & -     & -     & -     & -     & -     \\

MCAR  \cite{mcar}      & 83.8  & 85.0  & 72.1  & 78.0  & 88.0  & 73.9  & 80.3                          & 88.1  & 65.5  & 75.1  & 91.0  & 66.3  & 76.7  \\

% RARL \cite{rarl}      & -     & -     & -     & -     & -     & -     & -                             & 78.8  & 57.2  & 66.2  & 84.0  & 61.6  & 71.1  \\

% RDAL \cite{RDAL}   & -     & -     & -     & -     & -     & -     & -                             & 79.1  & 58.7  & 67.4  & 84.0  & 63.0  & 72.0  \\

SSGRL  \cite{SSGRL}     & 83.8  & \textbf{89.9}  & 68.5  & 76.8  & \textbf{91.3}  & 70.8  & 79.7    & \textbf{91.9}  & 62.5  & 72.7  & \textbf{93.8}  & 64.1  & 76.2  \\

% SRN    \cite{SRN}     & 77.1  & 81.6  & 65.4  & 71.2  & 82.7  & 69.9  & 75.8                          & 85.2  & 58.8  & 67.4  & 87.4  & 62.5  & 72.9  \\

% C-Trans \cite{C-Trans}    & 85.1  & 86.3  & 74.3  & 79.9  & 87.7  & 76.5  & 81.7                          & 90.1  & 65.7  & 76.0  & 92.1  & 71.4  & 77.6  \\
ADD-GCN \cite{ADD-GCN}    & 85.2  & 84.7  & 75.9  & 80.1  & 84.9  & 79.4  & 82.0  
                    & 88.8  & 66.2  & 75.8  & 90.3  & 68.5  & 77.9  \\
\hline\hline
Baseline     & 85.1  & 86.8  & 73.5  & 79.6  & 87.6  & 76.9  & 81.9                          & 90.1  & 65.2  & 75.7  & 91.7  & 67.9  & 78.0  \\
CSRA$^{*}$     & 85.2  & 87.5  & 73.0  & 79.6  & 88.1  & 75.7  & 81.5                          & 90.9  & 65.4  & 76.0  & 92.0  & 67.2  & 77.6  \\
Ours        & \textbf{86.4} & 84.8  &\textbf{77.4}  &\textbf{80.9}  & 85.4  &\textbf{80.3}  & \textbf{82.8}  & 89.5  & \textbf{67.5}  & \textbf{77.0}  & 91.2  & \textbf{69.6}  & \textbf{79.0}  \\ \hline      
\end{tabular}
\end{table*}

\subsection{Implementation details}
We load the model weights with the officially provided ViT-B\_16 model pre-trained on ImageNet21k \cite{ImageNet}. In all experiments, we use the SGD optimizer to optimize with a learning rate of 0.05. Momentum and weight decay are set to 0.9 and 1e-5, respectively. The cosine decay method is used to adjust the learning rate with a batch size of 32. COCO and VOC are trained for 10,000 steps and 2,000 steps, respectively, where warm-up steps are set to 5\% of all steps. In the training phase, we use the data augmentation suggested in \cite{ADD-GCN, ML-GCN} to avoid overfitting: the image is firstly resized to $512 \times 512$, then it is randomly cropped and resized to 448 × 448 with a random horizontal flip. In the test phase, we directly resize the image to 448*448. We cut the image into patches as in the ViT \cite{ViT}, with the patch size $16 \times 16$. Note that Unless otherwise stated, the default hyperparameters setting in all experiments is $TopN=4$, with the $\lambda$ 0.6. The ablation experiments provide a detailed analysis of the effects of each module and the hyperparameters. 
% We use Pytorch to build the entire framework and run all experiments on Tesla V-100 with 32G memory.

\subsection{Evaluation metrics}
To make a fair comparison with existing methods, we refer to the evaluation criteria from previous work \cite{RDAL, Asymmetric}. For the COCO dataset, we compare the mean accuracy (mAP), and the overall accuracy (OP). For the PASCAL-VOC dataset, we compare the average precision (mAP), and overall precision (OP), recall (OR), F1-measure (OF1) and per-class precision (CP), recall (CR), F1-measure (CF1) for further comparison. For the VOC dataset, we focus on comparing the average precision per category (AP) and the mean average precision across all categories (mAP).

\subsection{Comparisons with state-of-the-arts methods}
\subsubsection{Performance on MS-COCO}
We follow the traditional convention of training the model on the training set and reporting the precision, recall, and F1-measure of the model on the validation set, containing Top 1 and Top 3 metrics.
First, the baseline of our method is to use the ViT-B\_16 model to downscale the patch token with $1 \times 1$ convolution and then feed it into the average pooling operation. Finally, the output classification result is obtained. For fair comparisons, we deliberately select a recent open-source excellent works CSRA, and implement its core modules from its open-source code. The core modules are added to our baseline model, which will be trained and tested under the same experimental setup.
We refer to our method and CSRA combined with baseline, as DiD and CSRA\*, respectively, in our experiments. Table 1 summarizes our results and compares them with the CNN-based SOTA methods as well as CSRA\*. 
For mAP, DiD outperforms all CNN-based methods. DiD achieves 1.3\%, 1.2\% improvements over baseline and CSRA\*, respectively. DiD consistently outperforms other SOTA methods in other less important metrics such as CR, CF1, OR, and OF1. To analyze the localization of DiD on images, we show in Figure 3 the instance-aware prediction proposals generated by the dynamic instance localization module.

\subsubsection{Performance on PASCAL VOC 2007 and VOC 2012}
To make a fair comparison with other SOTA methods, we train our model on the training-validation set and evaluate it on the test set, as in previous work. Table 2 summarizes our results of DiD method. It can be seen that DiD achieves the best mAP results among all methods including the current SOTA method ADD-GCN. It outperforms ADD-GCN \cite{ADD-GCN} by 0.2\% (96.2\% vs. 96.0\%). Unlike VOC 2007, each method performing on VOC 2012 is required to upload the test outputs to its evaluation server to obtain the test results, which makes the results fairer. Table 2 reports the mAPs covering all categories. Our DiD also achieved the best performance compared to other SOTA methods. Specifically, DiD achieves an mAP value of 96.8\%, which is 1.3\% higher than the other SOTA.

% \subsubsection{Performance on PASCAL VOC 2012}
% To make a fair comparison with other methods, we train our model on the training-validation set and evaluate it on the test set, as in previous work. Unlike VOC 2007, each method performing on VOC 2012 is required to upload the test outputs to its evaluation server to obtain the test results, which makes the results fairer. Table 2 reports the mAPs covering all categories. Our DiD also achieved the best performance compared to other SOTA methods. Specifically, DiD achieves an mAP value of 96.8\%, which is 1.3\% higher than the other SOTA.
\subsection{Ablation Study}
We conduct ablation experiments on our DiD framework to analyze how its variants affect MLIR. All ablation studies are done on the COCO dataset.
    \textbf{Contribution of re-constraint strategy.}
We first analyze the contribution of our proposed re-constraint strategy to the overall framework and demonstrate its effectiveness. The first group is the baseline method (aka ViT-B\_16). The second is the first re-constraint strategy, i.e., the Hadamard product-based approach, the third group is the second re-constraint strategy, i.e., the element-wise summation approach, and finally is the third strategy, i.e., without introducing any spatial constraint relations. Table 3 reports all the experimental results, from which we can see that all three of our proposed re-constraint strategies have a large improvement on the baseline, with the first re-constraint strategy achieving an improvement of 1.9\%.

\begin{figure}[htb]
    \centering
    \subfigure[Fix $\lambda$=0.6]{
        \includegraphics[width=1.5in]{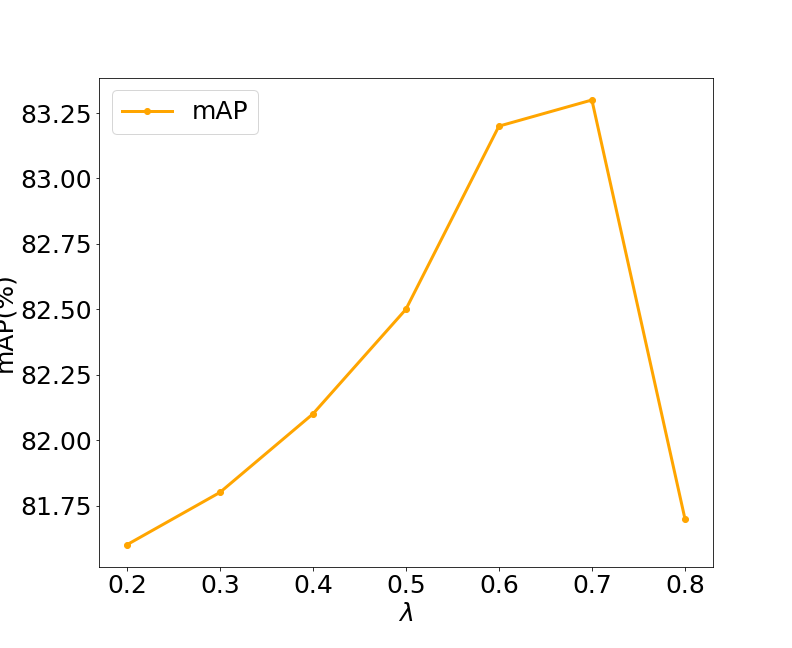}
    }
    \subfigure[Fix topN=4]{
	\includegraphics[width=1.5in]{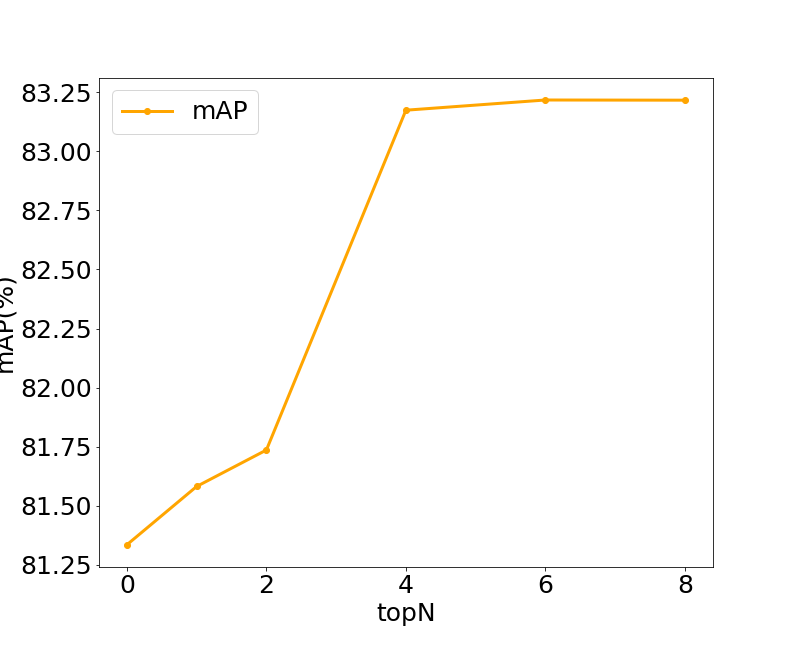}
    }
    \caption{The influence of $\lambda$ and topN on the COCO dataset. We set the input size to $256\times256$.}
    \label{fig.1}
\end{figure}

\begin{table}[tb]
\footnotesize
\centering
\caption{Comparison of our DiD and other state-of-the-art methods on Pascal VOC 2007 and Pascal VOC 2012.}
\begin{tabular}{l|l|l}
\hline
\multicolumn{1}{l}{} & \multicolumn{1}{l}{VOC 2007}  & \multicolumn{1}{l}{VOC 2012}   \\ \hline
Methods  & mAP  & mAP   \\ \hline\hline
HCP \cite{HCP}  & 90.9  & 90.5  \\
SSGRL \cite{SSGRL} & 95.0  & 94.8   \\
ADD-GCN \cite{ADD-GCN} & 96.0  & 95.5  \\
\hline\hline
Ours        & \textbf{96.2} & \textbf{96.8}  \\ \hline      
\end{tabular}
\end{table}

    \textbf{Effect of $\lambda$.}
Since the dynamic instance localization module in our method requires the choice of hyperparameter $\lambda$, we experimentally investigate the effect of $\lambda$ on the mAP value. We fix the number of topN to 4 and the input resolution to $256\times256$. We conduct 6 sets of $\lambda$ values between 0.2 and 0.8, with 0.1 interval. The results of all experiments are shown in the left subfigure of Figure 2. Note that when $\lambda$ equals 0, it is the baseline for our experiment. It can be seen that as the value of $\lambda$ increases from 0.2 to 0.8, the mAP rises and then falls. The best performance is taken when $\lambda$ takes 0.7. When $\lambda$ is greater than 0, the experimental results are always better than baseline. In this regard, we can analyze it this way: when $\lambda$ is small, the dynamic instance localization module will include as much global information from the raw image as possible, resulting in many non-critical areas being fed into the model again,  when $\lambda$ is large, the dynamic instance localization module will include as little of the raw image as possible, losing many critical areas to some extent.

    \textbf{Number of local instances.}
We experimentally investigate the effect of selecting local instances, that is, the value of topN, on the mAP value. We fix $\lambda$ to 0.6 and the input resolution to $256\times256$. We take topN values in the set $\{0,1,2,4,6,8\}$ for our experiments, and all the results are shown in the right subfigure of Figure 2. Note that when topN equals 0, it is the baseline for our experiment. It can be seen that the mAP performance tends to increase as the number of topN gradually increases. This implies that it is useful to utilize more local instances to improve the MLIR performance. The mAP tends to be stabie when the topN value is greater than 4, which corresponds to an average of 2.9 targets per image in the COCO dataset.

\begin{table}[tb]
\footnotesize
\centering
\caption{Comparison of mAP in \% of different re-constraint strategies on the COCO dataset.}
\begin{tabular}{l|l}
\hline
Methods                & mAP(\%) \\ \hline\hline
Baseline               & 81.3    \\
Hamad product-based    & 83.2    \\
Element-wise summation & 82.0   \\ 
Without constrain & 82.5   \\ \hline
\end{tabular}
\end{table}

\begin{table}[tb]
\footnotesize
  \begin{minipage}{0.45\linewidth}
    \centering
    \caption{Ablation Study of different instance localization strategies.}
    \begin{tabular}{l|l}
         \hline
Methods                & mAP(\%) \\ \hline\hline
ours               & \textbf{83.2}    \\
MCAR    & 81.6    \\
random & 81.6   \\ \hline
\end{tabular}
    \label{tab:mytab4}
  \end{minipage}
  \quad
  \begin{minipage}{0.45\linewidth}
    \centering
    \caption{Ablation Study of different instance localization strategies.}
        \begin{tabular}{l|l}
             \hline
Methods                & mAP(\%) \\ \hline\hline
Descending               & \textbf{83.2}    \\
Ascending   & 81.6    \\
Random & 81.6   \\ \hline
        \end{tabular}
    \label{tab:mytab5}
  \end{minipage}
\end{table}

% \subsubsection{Influence of image resolution}
% Image resolution is a critical factor for MLIR, which is very much reflected in real industrial scenarios. In order to investigate the performance improvements of our DiD for different image resolutions, we conduct 6 different pair of image resolutions for comparison experiments, which are $192 \times 192$, $256 \times 256$, $320 \times 320$, $384 \times 384$, and $448 \times 448$. Figure 3 reports the results of all experiments. It can be seen that DiD can have an average improvement of 1.6\% over the baseline. When the resolution is at $256 \times 256$, the improvement reaches 1.9\%.
    \textbf{Localization strategy.}
To validate the effectiveness of our proposed dynamic instance localization strategy, we design two different sets of localization strategies for comparison, the first one is the MCAR method, which obtains the prediction proposals based on the feature map peaks, and the second one is a method to obtain the prediction proposals randomly. The setup is the same for all experiments, and no new model parameters are introduced. From Table 4, it can be seen that our proposed localization strategy works significantly better than the two compared methods, demonstrating the effectiveness of the weakly supervised object localization-based approach.

\begin{figure}[tb]
\centering
\includegraphics[width=1.0\columnwidth]{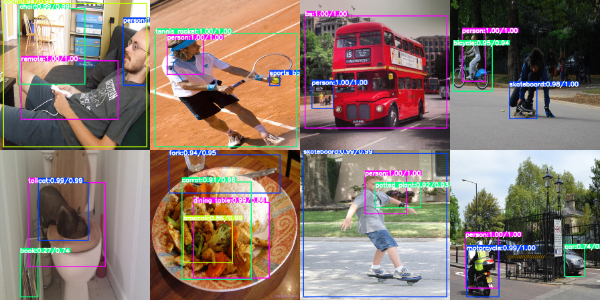} % Reduce the figure size so that it is slightly narrower than the column. Don't use precise values for figure width.This setup will avoid overfull boxes.
\caption{Selected examples of instance localization on MS-COCO 2014. For each sample, one color and value represent one object category and output probability, respectively. (Best view in color)}
\end{figure}

    \textbf{Descending vs ascending vs random.}
We analyze the contribution of the semantic feature maps selection strategy to the overall framework and demonstrate its effectiveness. We design two different selection strategies, the first one is to rank the semantic category output confidence in ascending order and then select topN feature maps, and the second one is to select N feature maps randomly. From Table 5, it can be seen that the descending order effect is significantly better than the two compared methods, which demonstrates the effectiveness of the semantic category-aware module.

\section{Conclusion}
In this work, we propose Diverse instance discovery (DiD): Vision-Transformer for instance-aware multi-label image recognition. Our core lies in the leverage of pure ViT to mine rich instances in multi-label images. To this end, Did first maps the patch tokens into category-aware representations by the semantic category-aware module, and then enhances the constraints of the spatial relations to these representations by spatial relationship-aware module and re-constraint strategy. Finally, the local instance region features from the raw images will be extracted by the dynamic instance localization module. Extensive experiments on public benchmarks demonstrate the effectiveness and rationality of our DiD. The future work is how to locate instance features more precisely to further improve the metrics.

\textbf{Acknowledgements} This work was supported by the NSFC projects (No. 62072399, No. U19B2042), Chinese Knowledge Center for Engineering Sciences and Technology, MoE Engineering Research Center of Digital Library, Alibaba-Zhejiang University Joint Institute of Frontier Technologies, and the Fundamental Research Funds for the Central Universities. 

% References should be produced using the bibtex program from suitable
% BiBTeX files (here: strings, refs, manuals). The IEEEbib.bst bibliography
% style file from IEEE produces unsorted bibliography list.
% -------------------------------------------------------------------------
\small
\bibliographystyle{IEEEbib}
\bibliography{DiD}

\end{document}